\documentclass[english]{ccdconf}
\usepackage{graphicx}
\usepackage{indentfirst} 
\usepackage{subcaption}
\usepackage{amsmath}
\usepackage{amssymb}
\usepackage{algorithm}
\usepackage{algorithmic}
\graphicspath{ {./images/} }
\begin{document}

\title{Search-Based Path Planning Algorithm for Autonomous Parking: Multi-Heuristic Hybrid A*}
\author{Jihao Huang\aref{zju}, Zhitao Liu\aref{zju}, Xuemin Chi\aref{zju}, Feng Hong\aref{zju}, Hongye Su\aref{zju}}

\affiliation[zju]{State Key Laboratory of Industrial, Control Technology, Zhejiang University, Hangzhou,310027
        \email{jihaoh@zju.edu.cn, ztliu@zju.edu.cn, chixuemin@zju.edu.cn, hogfeg.zju.edu.cn, hysu@iipc.zju.edu.cn}}

\maketitle

\begin{abstract}
This paper proposed a novel method for autonomous parking. Autonomous parking has received a lot of attention because of its convenience, but due to the complex environment and the non-holonomic constraints of vehicle, it is difficult to get a collision-free and feasible path in a short time. To solve this problem, this paper introduced a novel algorithm called Multi-Heuristic Hybrid A* (MHHA*) which incorporates the characteristic of Multi-Heuristic A* and Hybrid A*. So it could provide the guarantee for completeness, the avoidance of local minimum and sub-optimality, and generate a feasible path in a short time. And this paper also proposed a new collision check method based on coordinate transformation which could improve the computational efficiency. The performance of the proposed method was compared with Hybrid A* in simulation experiments and its superiority has been proved.
\end{abstract}


\footnotetext{This work was partially supported by National Key R\&D Program of China (Grant NO. 2021YFB3301000); Science Fund for Creative Research Group of the National Natural Science Foundation of China (Grant NO.61621002), National Natural Science Foundation of China (NSFC:62173297), Zhejiang Key R\&D Program (Grant NO. 2021C01198,2022C01035).}

\setlength{\parindent}{2ex} 
\section{INTRODUCTION}

Autonomous driving has received a lot of attention due to its various application scenarios \cite{ref11}, in which autonomous parking is a key problem in this area. Autonomous parking is a sub problem of path planning. Path planning means if the start position and goal position are given, a feasible and collision-free path can be found or reporting there is no solution. Path planning is a complex problem because it not only needs to comply with the constraints of the environment such as obstacles and traffic rules but also needs to guarantee the feasibility of the path, it is a high dimensional problem. 

Many path planning algorithms are proposed and they are classified into four groups according to their implementation: Graph search, Sampling, Interpolating and Numerical optimization \cite{ref1}. Algorithms based on Graph search are also called orderly sampling-based algorithms and the Sampling algorithms also refer to the random sampling-based algorithms. Different algorithms have different features, in actual use, we often combine different algorithms to get better characteristics, such as the combination of Rapid-Exploring Random Tree* (RRT*) and optimization method \cite{ref3}. Zhang and Liniger \cite{ref2} proposed a novel method that combines Hybrid A* and the optimization-based collision avoidance (OBCA) algorithm and called this method Hierarchical OBCA (H-OBCA), the main idea of H-OBCA is to use a global path planner to generate a coarse path and then use the coarse path as the initial guess of OBCA, then the OBCA algorithm can generate solution which could meet the dynamics of the vehicle. The solution quality of OBCA depends heavily on the quality of the initial guess, so it is necessary to get a better initial solution.

The algorithm belongs to graph search and sampling category are often used to generate global path solutions, such as A* and its variants, Rapid-Exploring Random Tree (RRT) and its variants, Probabilistic Roadmap (PRM), and other methods. But the algorithm based on graph search tends to be more efficient than the algorithm based on random sampling in states spaces whose dimensions are less than six\cite{ref4}. When using RRT and its variants in the path planning of Unmanned Aerial Vehicle (UAV), its performance will be better. So algorithms based on graph search are more popular in the path planning of autonomous vehicles. A* is an algorithm based on heuristic and it has many variants such as Dynamic A*, Hybrid A*, Weighted A*, and so on. Hybrid A* \cite{ref5} has been widely used in path planning for an autonomous vehicle because of its good quality, but in complex scenarios, it often costs much time to get a feasible solution through Hybrid A*. Multi-Heuristic A* (MHA*) \cite{ref6}  is a recently proposed algorithm that uses many heuristics to speed up the search process. It uses many inadmissible heuristics so it could make full use of the information provided by these heuristics to avoid local minimum in the search process. And MHA* has the ability to guarantee the completeness and sub-optimality of the solution. In \cite{ref7}, a method which incorporated the characteristic of MHA* and changed the extension manner of A* to make the path more feasible.

In this paper, we propose a method that combines MHA* and Hybrid A* called Multi-Heuristic Hybrid A* (MHHA*). The article is organized as follows. Section 2 introduces the problem formulation. Section 3 introduces the whole implementation of MHHA* and analyzes the effect of heuristic functions. Section 4 shows the simulation results of this work and makes a comparison with Hybrid A*. Conclusions are drawn in Section 5.

\section{PROBLEM FORMULATION}

In this section, we assume that all the scenario information can be obtained by the sensors without some errors creeping in. Then if the start position and the goal position are given, the autonomous parking algorithm could plan a path that could meet the non-holonomic constraints of the vehicle and does not collide with the environment, or report the non-existence of such a path. This work mainly focuses on the improvement of the classical path planner and we will introduce some necessary information for this problem.

\subsection{Environment}
\begin{figure}[h]
  \centering
  \includegraphics[width=\hsize]{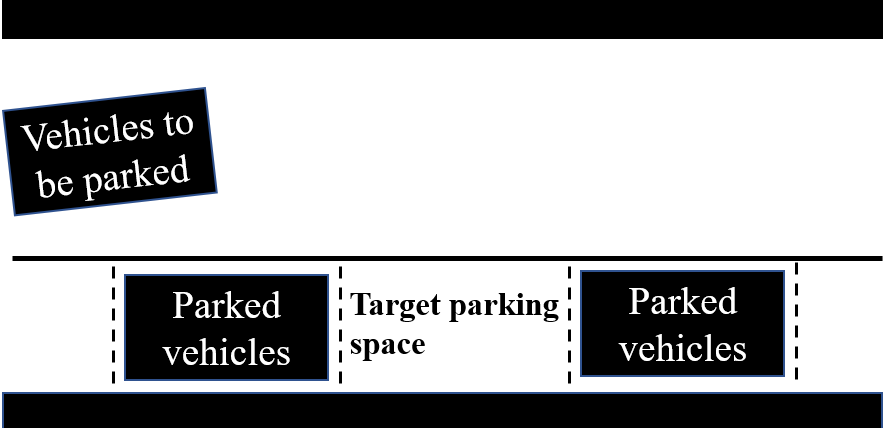}
  \caption{The parallel parking scenario.}
  \label{fig1}
\end{figure}

The parking problem is considered as one of the most common driving behaviors. Common parking scenarios include parallel parking, reverse parking, and oblique parking. To be honest, different parking arrangements could be solved through this algorithm, in this work we choose parallel parking as the research target. Figure \ref{fig1} could represent this scenario.

\subsection{Vehicle Kinematic Constraints}
\begin{figure}[h]
  \centering
  \includegraphics[width=\hsize]{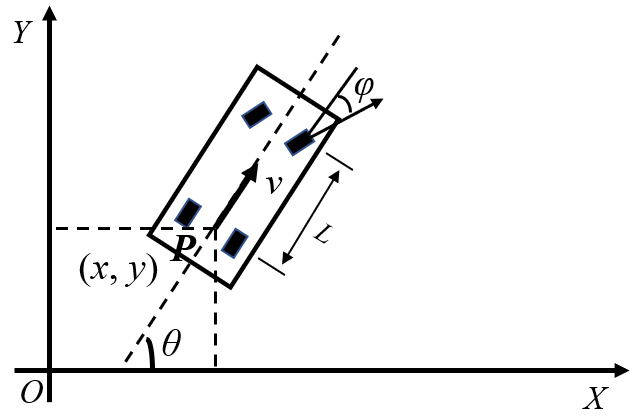}
  \caption{The motion model of vehicle.}
  \label{fig2}
\end{figure}

This subsection main focuses on the kinematic constraints of the vehicle and the vehicle kinematics model can be shown in Figure \ref{fig2}. As is known to us, driving in the unstructured scenario such as parking, the vehicle usually drives at a low velocity, so it is reasonable for us to simplify the vehicle model without thinking about the dynamics constraints. The kinematic model of vehicle can be formulated by $\dot{z}(t)=f(z(t),u(t))$ and we can use the single-track model to represent it \cite{ref8}:

\begin{equation}
 \label{eq2}
 \frac{d}{dt}
 \begin{bmatrix}
   x(t)\\
   y(t)\\
   \nu(t)\\
   \varphi(t)\\
   \theta(t)\\
 \end{bmatrix} = 
 \begin{bmatrix}
   \nu(t)\cdot\cos\theta(t)\\
   \nu(t)\cdot\sin\theta(t)\\
   a(t)\\
   \omega(t)\\
   \nu(t)\cdot\tan\varphi(t)/L
 \end{bmatrix}
\end{equation}
where $L$ is the wheelbase, ($x$, $y$) is the rear-wheel axle mid-point (point $P$ in Figure \ref{fig2}), $\nu$ is the velocity of $P$, $\varphi$ is the  steering angle, $\theta$ is the heading direction, $\omega$ is the angular velocity of the front wheel, $a$ is the acceleration. So we can reformulate the equation $\dot{z(t)}=f(z(t),u(t))$ as $z(t) = [x(t), y(t), \nu(t), \varphi(t), \theta(t)]$, $u(t) = [a(t), \omega(t)]$. 

The state/control variables mentioned earlier have a range, which reflects the physical or mechanical limitations of the vehicle, it can be represented by:
\begin{equation}
 \begin{bmatrix}
   a_{min}\\
   \nu_{min}\\
   -\omega_{max}\\
   -\varphi_{max}
 \end{bmatrix}\leq
 \begin{bmatrix}
   a(t)\\
   \nu(t)\\
   \omega(t)\\
   \varphi(t)
 \end{bmatrix}\leq 
 \begin{bmatrix}
   a_{max}\\
   \nu_{max}\\
   \omega_{max}\\
   \varphi_{max}
 \end{bmatrix}
\end{equation}

\subsection{Collision Check}
This section mainly focused on how to avoid collisions with obstacles in the environment. We introduce coordinate transformation to simplify the collision detection process, which will improve computational efficiency greatly. The geometry of the vehicle in this work is approximated as a rectangle shape. We will do a collision check with two steps. Firstly, the car could be reasonably approximated by overlapping circular disks, a rectangle with length $l$ and width $\omega$ can be covered by n circular disks of radius $r$ calculated as \cite{ref9}:
\begin{equation}
 r=\sqrt{\frac{l^2}{n^2}+\frac{\omega^2}{4}}
\end{equation}

And the distance between circular disks can be calculated as:
\begin{equation}
 d=2\sqrt{r^2-\frac{\omega^2}{4}}
\end{equation}

\begin{figure}
  \centering
  \includegraphics[width=\hsize]{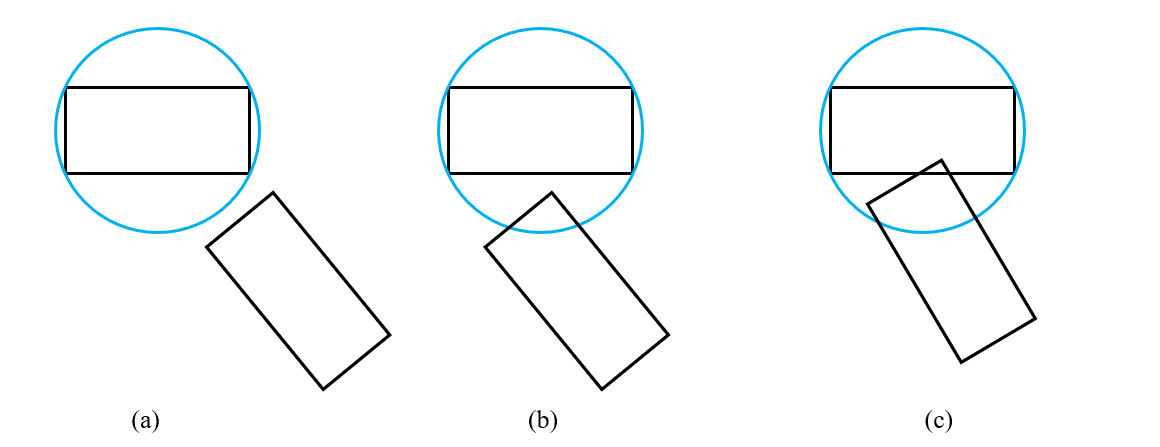}
  \caption{Examples for coarse collision check.}
  \label{fig3}
\end{figure}

In this work, we use one circular disk to cover the rectangle shape, as it can be shown in Figure \ref{fig3}. We can do a coarse check this way. When the distance between the center of the rectangle and the point on the obstacle is greater than the radius of the circular disk, we can make sure that the vehicle won't collide with the obstacle, it can be shown as Figure \ref{fig3} (a). But when the distance is less than the radius of the disk, the vehicle has the possibility of collision, as it can be shown in Figure \ref{fig3} (b), (c). So we need the next step.

\begin{figure}[h]
  \centering
  \includegraphics[width=\hsize]{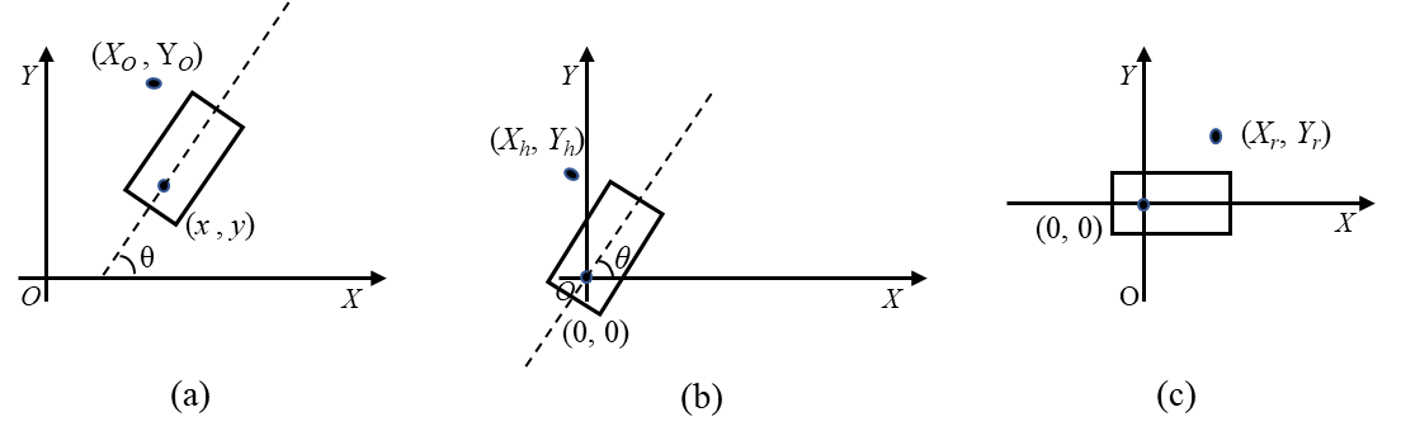}
  \caption{Complete collision check through coordinate transformation.(a):origin coordinate system.(b):horizontal transformation.(c):rotation transformation.}
  \label{fig4}
\end{figure}

We need to determine whether the point is inside the rectangle in order to overcome the defect shown in Figure \ref{fig3} (b). If we only check collision through the first step, even if there is a solution, we may not get a solution because of the waste of space. We use coordinate transformation to cope with this problem, it can improve computational efficiency greatly. As we can see in Figure \ref{fig4} (a), this is the earth coordinate system, ($x$, $y$) is the coordinate of the rear-wheel axle mid-point in the earth coordinate system, ($X_O$, $Y_O$) is the coordinate of the point on the obstacle in the earth coordinate system, $\theta$ is the orientation angle. The coordinate transformation consists of two parts, horizontal transformation and rotation transformation. In Figure \ref{fig4} (b), we establish a coordinate system which is obtained by horizontal transformation of the earth coordinate system, this coordinate system uses the rear-wheel axle mid-point as the coordinate origin, ($X_h$,$Y_h$) is the coordinate of the point on obstacle after horizontal transformation, we can calculate it as:
\begin{equation}
 \begin{bmatrix}
   X_h\\
   Y_h
 \end{bmatrix}=
 \begin{bmatrix}
   X_O\\
   Y_O
 \end{bmatrix}-
 \begin{bmatrix}
   x\\
   y
 \end{bmatrix}
\end{equation}

Then we establish a coordinate system that uses the rear-wheel axle mid-point as the coordinate origin and its horizon axis is parallel to the vehicle, as it can be shown in Figure \ref{fig4} (c). So if we can get the coordinate of the point on the obstacle based on this coordinate system, it will be easy for us to judge whether the point is inside the rectangle.($X_r$, $Y_r$) is the coordinate of the point on the obstacle in the coordinate system shown in Figure \ref{fig4} (c), we can calculate it as:
\begin{equation}
 \begin{bmatrix}
   X_r\\
   Y_r
 \end{bmatrix}=
  \begin{bmatrix}
   \cos(-\theta) & -\sin(-\theta)\\
   \sin(-\theta) &  \cos(-\theta)\\
 \end{bmatrix}
 \begin{bmatrix}
   X_h\\
   Y_h
 \end{bmatrix} 
\end{equation}

So we can get the coordinate of the point on the obstacle in coordinate system shown in Figure \ref{fig4} (c) through horizontal transformation and rotation transformation from the coordinate in the earth coordinate system, we can summarize it as:
\begin{equation}
 \centering
 x_h=Rx_o+t
\end{equation}
where $x_h$ represents the coordinate in the transformed coordinate system, R represents a rotation matrix, $x_o$ represents the coordinate in the earth coordinate system, t represents the translation vector.
\section{ALGORITHM}
This section mainly focused on introducing the Multi-Heuristic Hybrid A* Algorithm(MHHA*), which is based on Hybrid A* while incorporating the character of the Multi-Heuristic A* (MHA*) Algorithm. We can assure the kinematic feasibility through the expanding manner of Hybrid A* and fast search efficiency through MHA*.
\subsection{Heuristic}
This subsection mainly focuses on analyzing the effect of the heuristic function and deciding the heuristic function used in the Multi Heuristic Hybrid A* Algorithm. The performance of path planners based on heuristic function depends heavily on the performance of heuristic function \cite{ref6}. Heuristic function is used to reduce the exploration of useless areas and speed up the search process, but only an admissible heuristic can lead to the optimal results. A heuristic function is admissible if it never overestimates the cost from the current state to the goal state, it means that the distance between the current state and the goal state calculated by the heuristic function always less than the actual distance between between the current state and the goal state \cite{ref10}. A heuristic function is consistent if it satisfies $h(x_{goal})=0$ and $h(s) \leq h(s^{'})+c(s,s^{'})$, it can be shown as Figure \ref{fig5}:

\begin{figure}[h]
    \centering
    \includegraphics[width=\hsize]{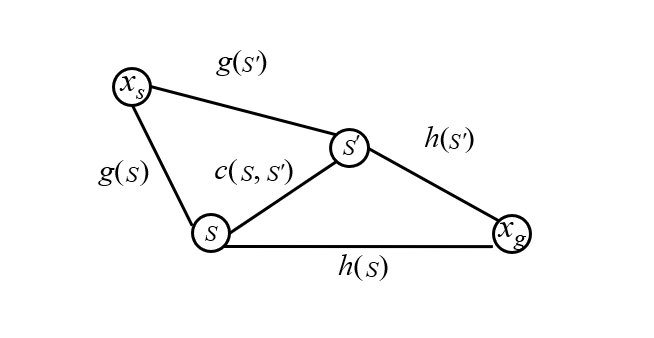}
    \caption{The consistent characteristic of the heuristic function.}
    \label{fig5}
\end{figure}
\noindent where $x_s$ represents the start state, $x_g$ represents the goal state, $S$ represents the current state, $S^{'}$ represents the child node of the current state, $g(s)$ represents the cost between $S$ and $x_s$, $h(s)$ represents the cost between $S$ and $x_g$, $c(s,s^{'})$ represents the cost between $S$ and $S^{'}$, so $g(s^{'})$ could be represented by $g(s^{'})=g(s)+c(s,s^{'})$ and the total cost $f(s)=g(s)+h(s)$. When the heuristic function underestimates the cost between the current state and the goal state, it means the cost is less than the actual cost, so the optimal solution can be found. When $h(s)=0$, then the algorithm degenerated into Dijkstra. When the heuristic function overestimates the cost between the current state and the goal state, the search speed will increase, but the solution may not be optimal, it means a trade-off is made between optimality and speed.

Two typical admissible heuristics are used, one takes into account the non-holonomic constraints of the vehicle while neglecting the obstacle and another considers the obstacle while neglecting the non-holonomic of the vehicle, we choose the maximum of both. For more detailed information about these heuristics, please refer to \cite{ref5}.

\subsection{Multi-Heuristic A* Algorithm}
This subsection mainly focuses on having a review of MHA*. MHA* uses many heuristics, one of those heuristics is admissible, while others are inadmissible. The admissible heuristic is also consistent, it
can assure the optimality and completeness of the solution but it has the defect that is easy to fall into local minimum when searching in the complex scenarios. Different inadmissible heuristics may be useful in different search places, which means different inadmissible heuristics could provide different guiding powers. MHA* combines the admissible heuristic and the inadmissible heuristics so it could get a complete and bounded sub-optimal solution. MHA* could assure fast convergence if one of these inadmissible heuristics can guide the search around the depression regions. MHA* runs multiple searches with different inadmissible heuristics in a manner that will assure sub-optimality. MHA* has two variants, including Independent Multi-Heuristic A*(IMHA*) and Shared Multi-Heuristic A*(SMHA*). For more details about MHA*, it is recommended to see the full paper \cite{ref6}.

\subsection{Multi-Heuristic Hybrid A* Algorithm}
This subsection mainly focuses on introducing the Multi-Heuristic Hybrid A* Algorithm, this algorithm incorporates the characteristic of Hybrid A* and the characteristic of MHA*, so it can assure the solution is kinematically feasible, complete, and sub-optimal.

MHHA* is presented in Alg \ref{alg3}. It will return the path connecting the start state and the goal state or report there's no collision-free solution connecting the start state and the goal state. The first step is to complete the initialization process(line1-10). Then this algorithm checks whether the $OPEN_{0}$ is an empty set, if it is an empty set, the algorithm will return no solution(line 40). Then the algorithm will expand the node. MHHA* uses $\omega$ factor to prioritize the inadmissible searches over the anchor search which is searched by the admissible heuristic functions. MHHA* executes the inadmissible searches in a round-robin manner as long as they explore solutions within $\omega$ factor of the minimum key of the anchor search. If the minimum key of an inadmissible heuristic cannot satisfy the bound of $\omega$ times of the minimum key of an anchor heuristic, this inadmissible search will be suspended and the anchor search will run instead. This is a process that determines which node will be expanded. After determining which node to be expanded, the process in lines 14-24 is the same as lines 26-36. Take the previous one as an example, first, the algorithm will check whether the current node to be expanded is the goal node, if it is, the algorithm will return the whole path through backtracking the parent node. If not, the algorithm tries to get an analytic solution, this algorithm chooses Reeds-Sheep curves because it allows the vehicle to move in both forward and backward directions. If the path returned by Reeds-Sheep curves is collision-free, then the algorithm will return the path which is combined with the Reeds-Sheep curves and backtracking the parent node. If the Reeds-Sheep curve collides with the obstacle, then the algorithm will call $Expandnode(s)$ function to finish the expansion of the current node. 

The $Expandnode(s)$ function is presented in Alg \ref{alg2}. This algorithm is similar to the expansion of A*, in addition to it not only using an admissible heuristic to expand the node as well as inadmissible heuristics. The $key(s,i)$ function that appeared in the algorithm is presented in Alg \ref{alg1}, which is used to calculate the total cost of one node.

\begin{algorithm}
\caption{Multi-Heuristic Hybrid A*}
\label{alg3}
\textbf{Function}: MHHA*($s_{start}$ , $s_{goal}$)
\begin{algorithmic}[1]
\STATE $g(s_{goal}) \leftarrow \infty$; 
\STATE $g(s_{start}) \leftarrow 0$; 
\STATE $bp(s_{start}) \leftarrow null$; 
\STATE $bp(s_{goal}) \leftarrow null$; 
\STATE $N \leftarrow 0$;
\FOR{$i=0$ to $n$}
\STATE $OPEN_{i} \leftarrow \emptyset$;
\STATE $CLOSE{i} \leftarrow \emptyset$;
\STATE insert $s^{start}$ in $OPEN_{i}$ with $key(s_{start},i)$ as priority;
\ENDFOR
\WHILE{$OPEN_{0}$ not empty}
\FOR{$i=1$ to $n$}
\IF{$OPEN_{i}.Minkey() \leq \omega OPEN_{0}.Minkey()$}

\STATE $N \leftarrow N+1$;
\IF{$g(s_{goal}) \leq OPEN_{i}.Minkey()$}
\RETURN path pointed by bp($s_{goal}$);
\ENDIF
\STATE $s \leftarrow OPEN_{i}.Top()$;
\IF{$mod(N, setvalue) == 0$}
\STATE RSpath $\leftarrow$ RSCurve($s,s_{goal}$);
\IF{RSpath is Collision free}
\RETURN RSpath with path pointed by bp(s)
\ENDIF
\ENDIF
\STATE Expandnode($s$);
\ELSE

\STATE $N \leftarrow N+1$;
\IF{$g(s_{goal}) \leq OPEN_{0}.Minkey()$}
\RETURN path pointed by bp($s_{goal}$);
\ENDIF
\STATE $s \leftarrow OPEN_{0}.Top()$;
\IF{$mod(N, setvalue) == 0$}
\STATE RSpath $\leftarrow$ RSCurve($s,s_{goal}$);
\IF{RSpath is Collision free}
\RETURN RSpath with path pointed by bp(s)
\ENDIF
\ENDIF
\STATE Expandnode($s$);

\ENDIF

\ENDFOR
\ENDWHILE

\RETURN NULL
\end{algorithmic}
\end{algorithm}

\begin{algorithm}
\caption{The Extension of Parent Node}
\label{alg2}
\textbf{Function}: Expandnode($s$)
\begin{algorithmic}[1]
\STATE Remove s from $OPEN_{i}$, $\forall i=0 \dots n$;
\STATE Add s to $CLOSE_{i}$, $\forall i=0 \dots n$;
\FOR{each $s^{'}$ in Succ(s) }
\IF{$s^{'}$ in $CLOSE_0$ }
    \STATE Continue;
\ELSE
    \STATE $g(s^{'}) \leftarrow g(s)+c(s,s^{'})$;
    \STATE $bp(s^{'}) \leftarrow s$;
    \STATE insert/update $s^{'}$ in $OPEN_{0}$ with $key(s^{'},0)$ as priority;
    \IF{$s^{'}$ not in any $CLOSE_i$ , $\forall i=1 \dots n$}
    \FOR{$i=1$ to $n$}
        \STATE insert/update $s^{'}$ in $OPEN_{i}$ with key($s^{'},i$);
    \ENDFOR
    \ENDIF
\ENDIF
\ENDFOR
\end{algorithmic}
\end{algorithm}

\begin{algorithm}
\caption{Calculate The Total Cost of One Node}
\label{alg1}
\textbf{Function}: key($s,i$)
\begin{algorithmic}[1]
\RETURN $g(s)+h_{i}(s)$
\end{algorithmic}
\end{algorithm}

\section{SIMULATION RESULTS}
In this section, we provide simulation results of MHHA* and compare the MHHA* with Hybrid A*.
\subsection{Simulation Setup}
\begin{figure}[h]
  \centering
  \includegraphics[width=\hsize]{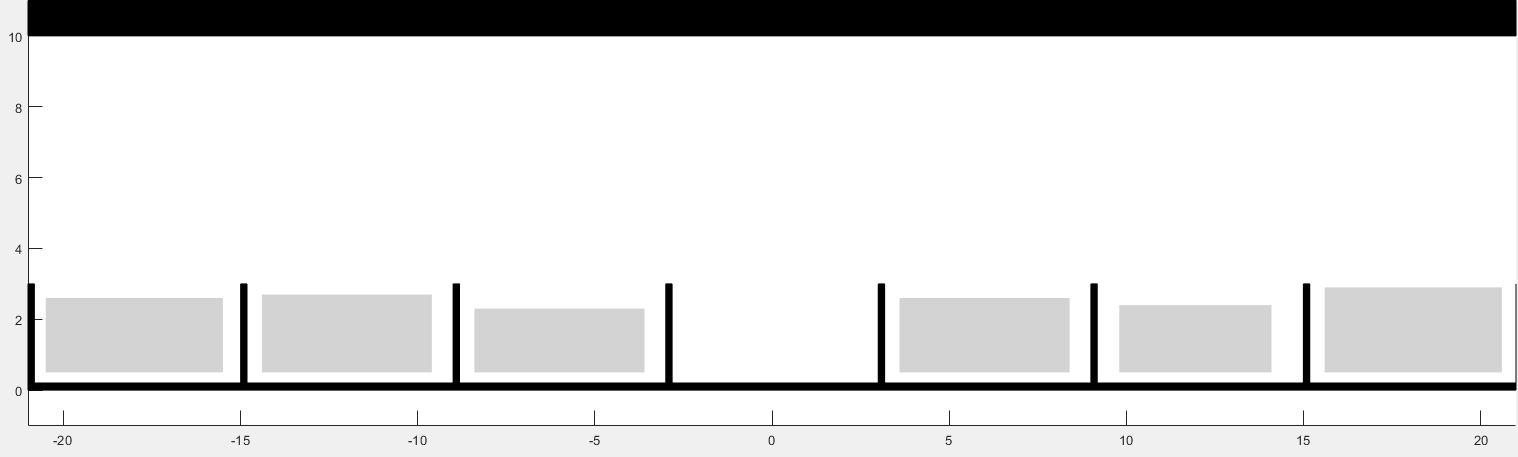}
  \caption{The experiment parking scenario.}
  \label{fig6}
\end{figure}

All the simulation results are conducted in MATLAB and executed on a laptop with i7-9750 CPU 2.60GHz with 16GB of RAM under Microsoft Windows 10. We model the vehicle as a rectangle of size $4.7\times2$m and the wheelbase L of the car is 2.7m. And in this work, the steering angle is limited between $\varphi \in [-0.6,0.6]$ rad (approximately $\pm 34^{\circ}$). This work choose parallel parking as research scenario, the horizontal direction is limited between $X \in [-21,21]$ m, another direction is limited between $Y \in [-1,11]$ m and the parking spot is 3.0 m deep and 7.2 m long, as is shown in Figure \ref{fig6}.

In MHHA* algorithm grid map is implemented and in both $X$ and $Y$ direction the size of the grid is 0.3 m. And this algorithm penalizes the reverse movement, numbers of switchbacks, and the jerk of the vehicle. For admissible heuristic, we choose the maximum of both, one takes into account the non-holonomic constraints of the vehicle while neglecting the obstacle and another considers the obstacle while neglecting the non-holonomic of the vehicle. For convenience, in this implementation, MHHA* only chooses one inadmissible heuristic and the inadmissible heuristic is obtained by inflating the admissible heuristic. 

\subsection{Results}
\begin{figure}[h]
  \centering
  \includegraphics[width=\hsize]{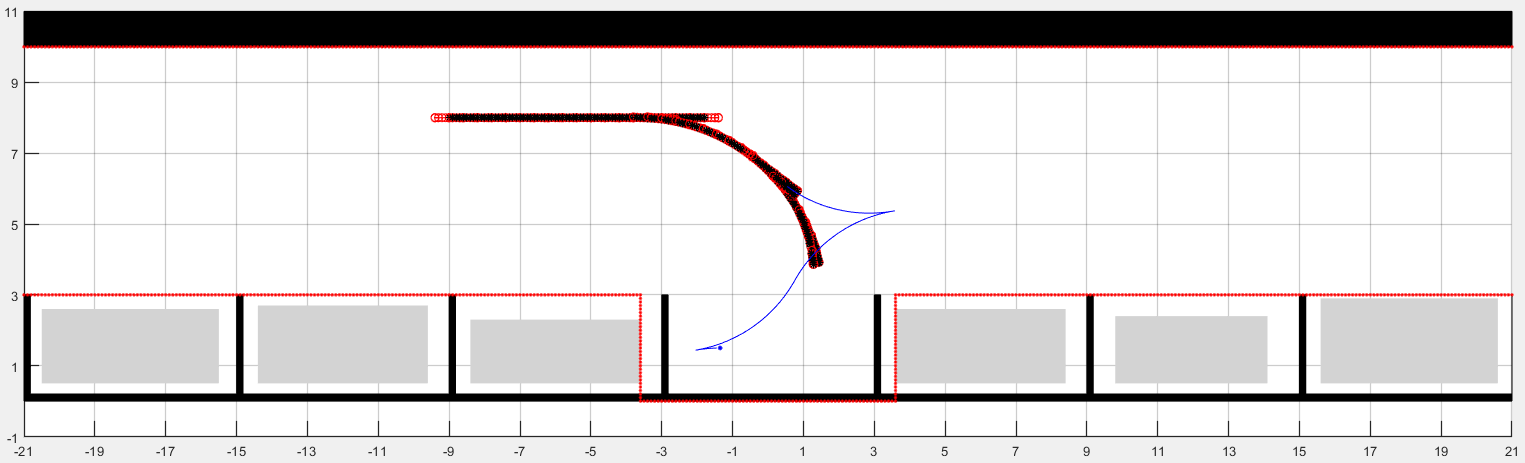}
  \includegraphics[width=\hsize]{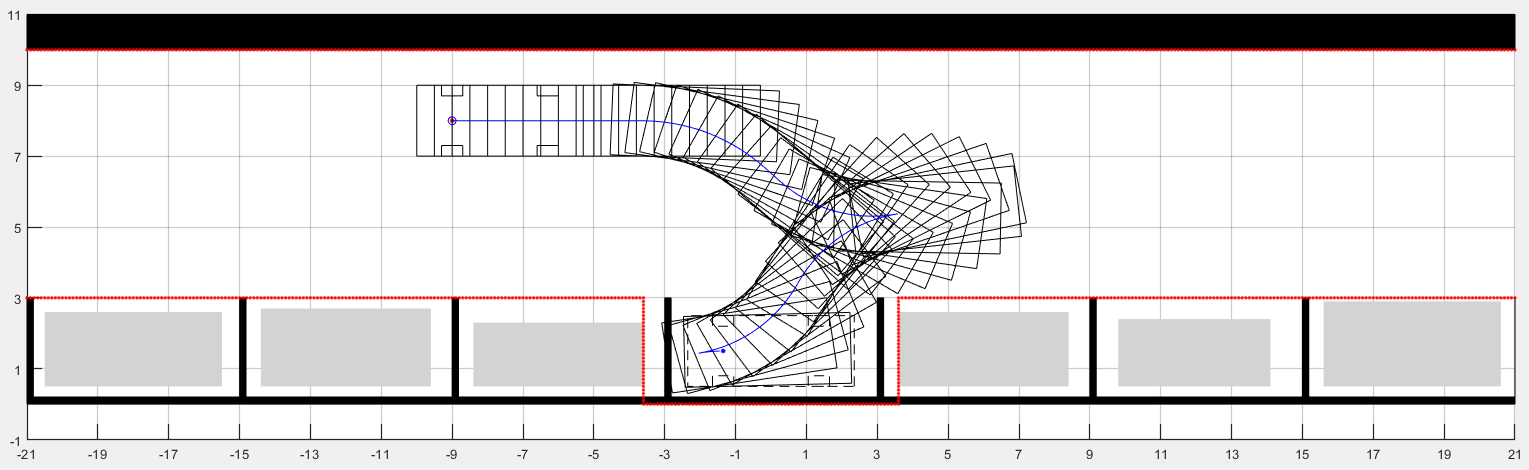}
  \caption{The forward parking of using MHHA*.}
  \label{fig7}
\end{figure}

\begin{figure}[h]
  \centering
  \includegraphics[width=\hsize]{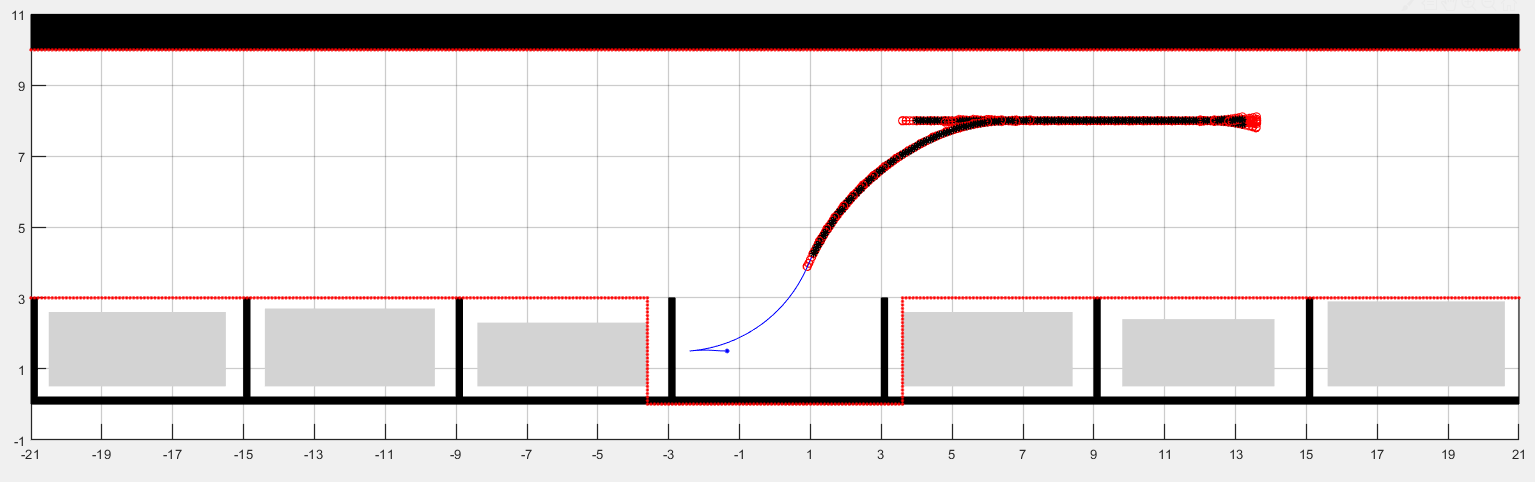}
  \includegraphics[width=\hsize]{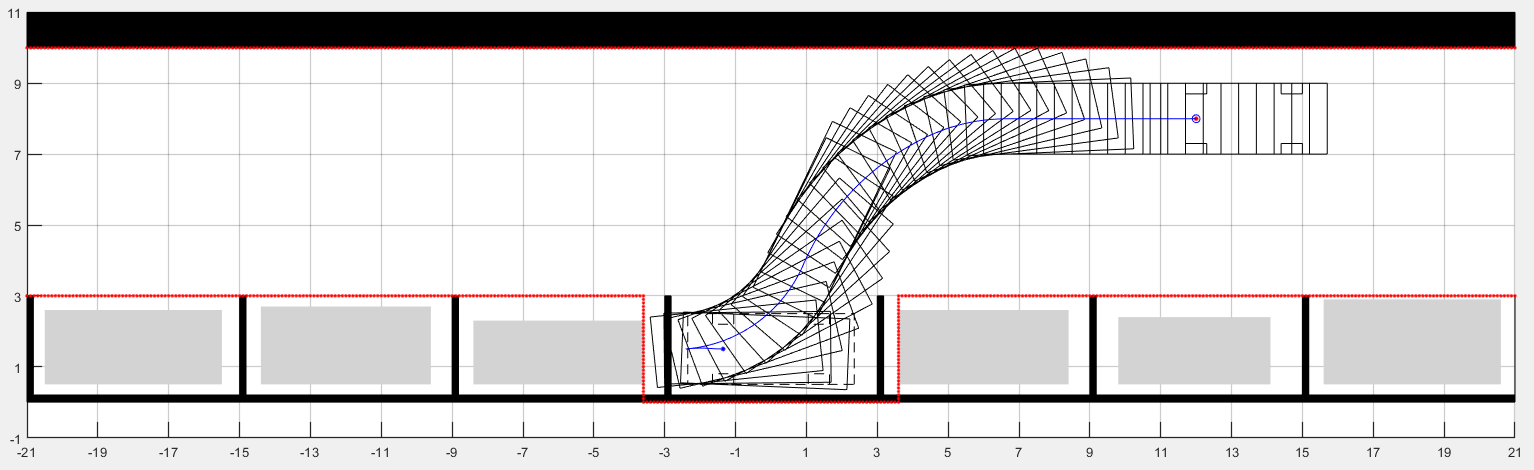}
  \caption{The backward parking of using MHHA*.}
  \label{fig8}
\end{figure}

In the simulation of this work,the end position is fixed at $(x_{f},y_{f},\varphi_{f},\nu_{f})=(-1.35,1.5,0,0)$, the start status is static ($\nu=0$) and the initial orientation is right. In this work two typical scenarios are selected to check the performance of MHHA*: one is forward entry parking, as shown as Figure \ref{fig7}, the other is backward entry parking, as shown as Figure \ref{fig8}. The obstacle is represented by the red points and the analytic solution is represented by the blue line. The black vertical line is not the obstacle, it is just used to represent the scope of the parking space.The start position in Figure \ref{fig7} is set in $(x_{s},y_{s},\varphi_{s},\nu_{s})=(-9,8.0,0,0)$ and the start position in Figure \ref{fig8} is set in
$(x_{s},y_{s},\varphi_{s},\nu_{s})=(12,8.0,0,0)$. The search process of MHHA* is very fast and this algorithm can get a collision-free solution in short time while assuring the sub-optimality.

\begin{figure}[h]
  \centering
  \includegraphics[width=\hsize]{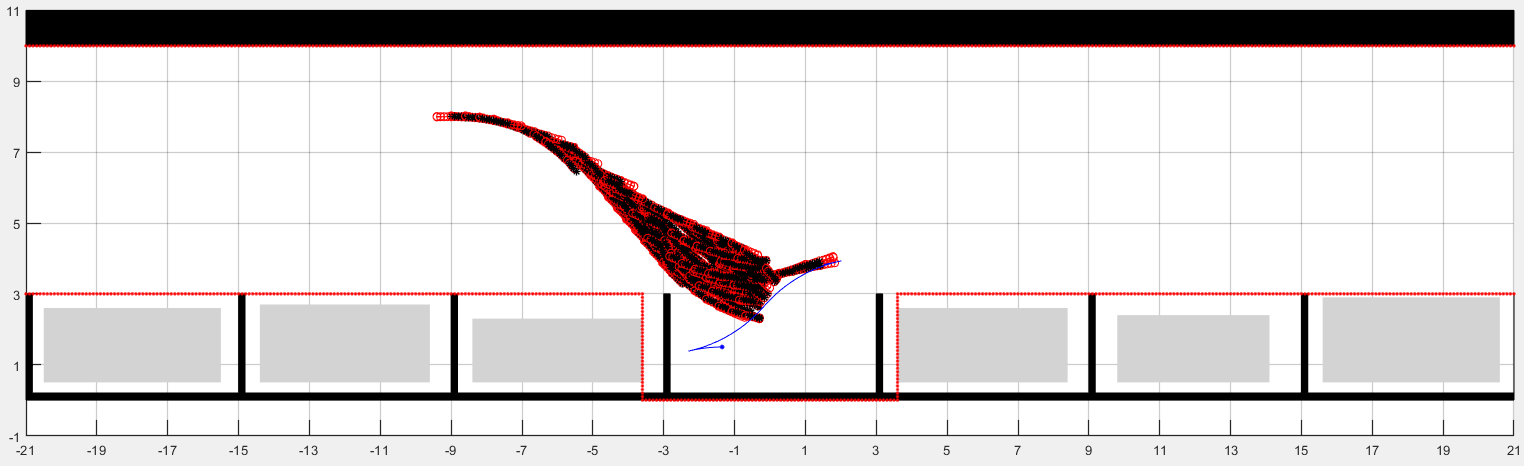}
  \includegraphics[width=\hsize]{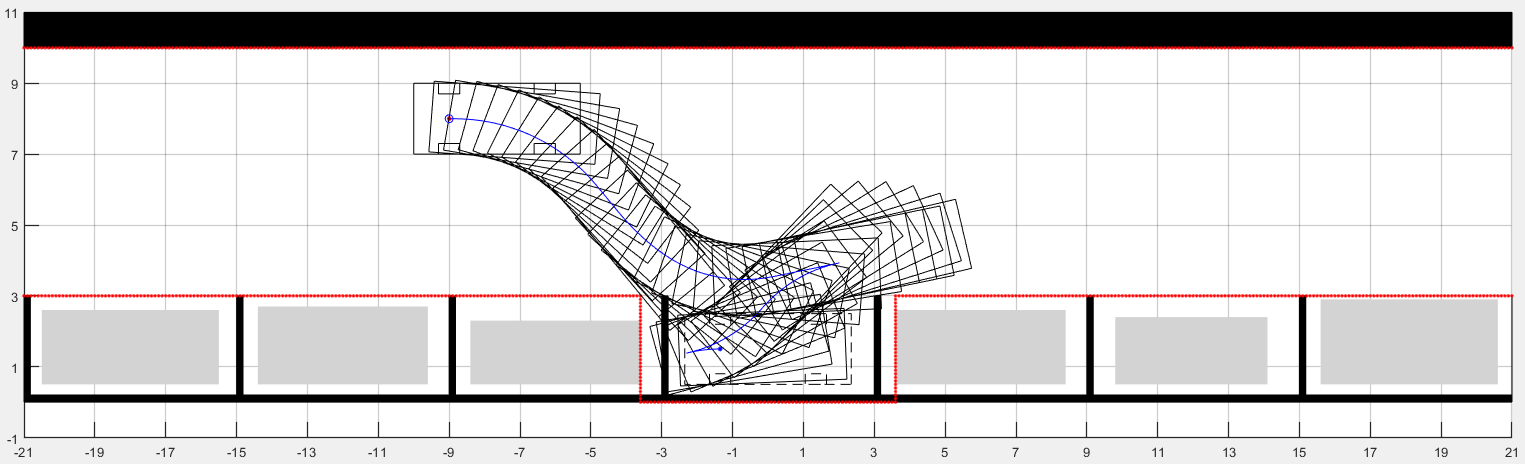}
  \caption{The forward parking of using Hybrid A*.}
  \label{fig9}
\end{figure}

\begin{figure}[h]
  \centering
  \includegraphics[width=\hsize]{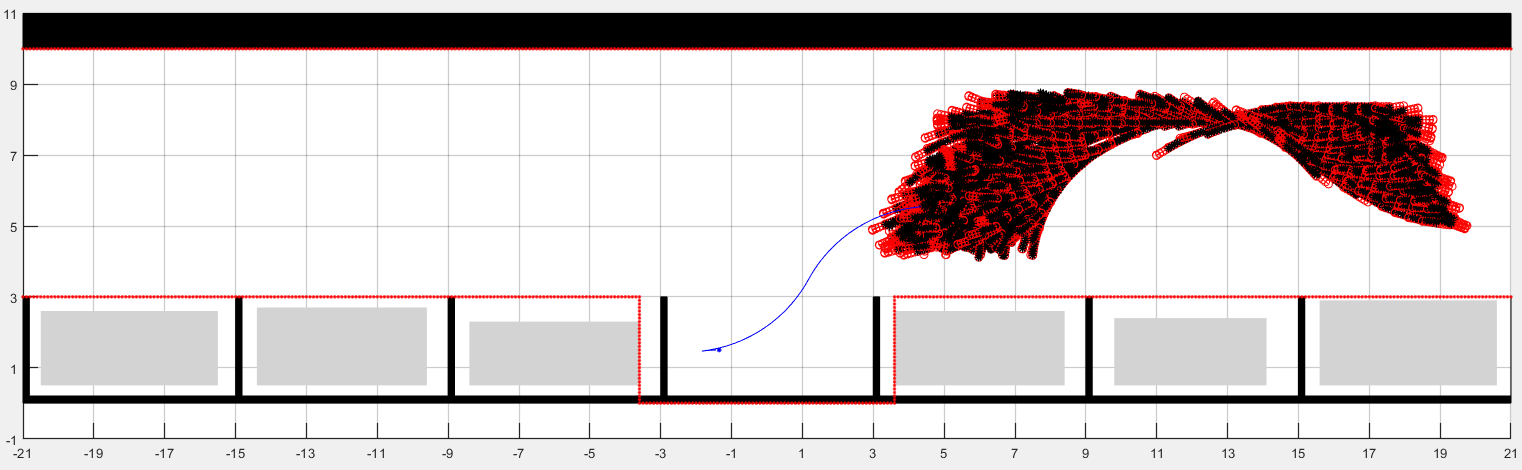}
  \includegraphics[width=\hsize]{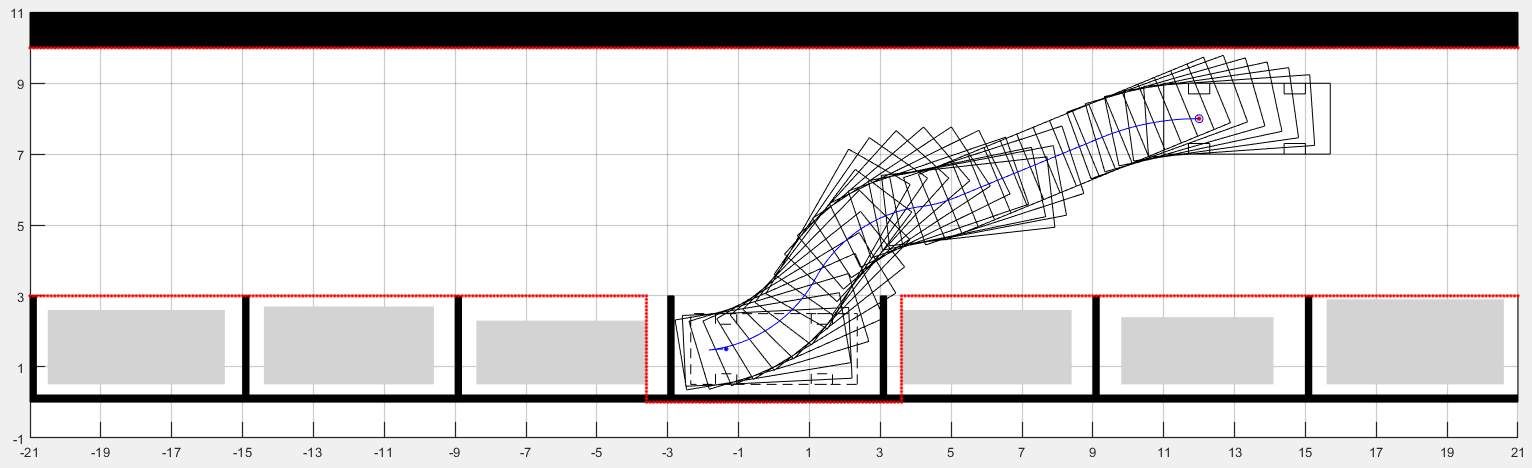}
  \caption{The backward parking of using Hybrid A*}
  \label{fig10}
\end{figure}

\subsection{Discussion and Comparison}
This subsection will compare the MHHA* with Hybrid A*. The simulation result of Hybrid A* in both scenarios can be shown as Figure \ref{fig9} and Figure \ref{fig10}. And the comparison of the key performance indicators between MHHA* and Hybrid A* is shown in Table \ref{tab1} and Table \ref{tab2}. We can see  MHHA* is better than Hybrid A* in many aspects, such as the number of extended nodes, the number of iterations, and the extension time. Only in the path length aspect the MHHA* is worse than Hybrid A*, because MHHA* makes a trade-off between optimality and computational efficiency, it only ensures sub-optimality. Although the path calculated by MHHA* is not optimal, it is feasible, so it could be used as the initial solution for OBCA. And the paths calculated by MHHA* and Hybrid A* are homotopy that is both solutions will converge to the same solution, so the computational efficiency is more important than optimality, while MHHA* is sub-optimal. So in the H-OBCA algorithm, we can choose MHHA* to generate the initial solution to improve the computational efficiency.
\begin{table}[h]
  \centering
  \caption{The performance comparison of forward parking}
  \label{tab1}
  \begin{tabular}{l|l|l}
    \hhline
    Performance               & MHHA*     & Hybrid A*  \\ \hline
    Number of Extended Nodes  & 273       & 1460       \\ \hline
    Number of Iterations      & 79        & 564        \\ \hline
    Extension Time (s)        & 1.0222    & 6.9123     \\ \hline
    Path lengths (m)          & 21.097    & 18.659     \\ 
    \hhline
  \end{tabular}
\end{table}

\begin{table}[h]
  \centering
  \caption{The performance comparison of backward parking}
  \label{tab2}
  \begin{tabular}{l|l|l}
    \hhline
    Performance               & MHHA*    & Hybrid A*    \\ \hline
    Number of Extended Nodes  & 253      & 6361         \\ \hline
    Number of Iterations      & 62       & 2486          \\ \hline
    Extension Time (s)        & 0.9753   & 43.49         \\ \hline
    Path lengths (m)          & 18.16321 & 16.691         \\ 
    \hhline
  \end{tabular}
\end{table}

\section{CONCLUSION}
This work proposed a search-based motion planning algorithm for autonomous parking: Multi-Heuristic Hybrid A* (MHHA*), it could provide the guarantee of sub-optimality and fast computational efficiency. And a  novel collision check method based on coordinate transformation is utilized to improve computational efficiency. Parallel parking is chosen as the research target, and the single-track model is selected as the vehicle model. In simulation experiments, two typical scenarios are selected, one is forward parking and the other is backward parking. Although MHHA* is not optimal, when used in OBCA as the initial solution, it will converge to the solution as same as Hybrid A* because of homotopy. Moreover, the MHHA* has good performance than Hybrid A* in the number of extended nodes, the number of iterations, and the extension time in both scenarios. Our future research direction is checking the performance of MHHA* in H-OBCA.

%\section*{ACKNOWLEDGEMENTS}

\bibliographystyle{IEEEtran}
\bibliography{reference}

% Note: place a \balance command somewhere within the left column of the
% last page will balance the two columns on the last page.
%
\balance

\end{document}